# A HIGH QUALITY TEXT-TO-SPEECH SYSTEM COMPOSED OF MULTIPLE NEURAL NETWORKS

*Orhan Karaali, Gerald Corrigan, Noel Massey, Corey Miller, Otto Schnurr and Andrew Mackie*

Speech Processing Laboratory
Motorola, Inc.
1301 E. Algonquin Rd., Schaumburg, IL 60196, USA

## ABSTRACT

While neural networks have been employed to handle several different text-to-speech tasks, ours is the first system to use neural networks throughout, for both linguistic and acoustic processing. We divide the text-to-speech task into three subtasks, a linguistic module mapping from text to a linguistic representation, an acoustic module mapping from the linguistic representation to speech, and a video module mapping from the linguistic representation to animated images. The linguistic module employs a letter-to-sound neural network and a postlexical neural network. The acoustic module employs a duration neural network and a phonetic neural network. The visual neural network is employed in parallel to the acoustic module to drive a talking head. The use of neural networks that can be retrained on the characteristics of different voices and languages affords our system a degree of adaptability and naturalness heretofore unavailable.

## 1. INTRODUCTION

We have developed a text-to-speech synthesizer including five cooperating neural networks, each specializing in a particular area of human natural language ability. This synthesizer represents a significant advance in naturalness and adaptability to different voice qualities and languages over previous rule-based and concatenative approaches.

Two linguistic neural networks in conjunction with a text preprocessor construct a linguistic representation including the pronunciation of incoming text. The linguistic representation is then passed to two acoustic neural networks which output speech parameters and a visual neural network which drives a talking head. These different neural network components feature modular architectures, time delay neural networks and recurrent internal data paths. Most of the networks are designed to be trained on a labeled speech database of a particular speaker. In this way, particular voice characteristics can be captured at each level of the system, resulting in a very natural voice quality.

The use of neural networks in the solution of these problems has helped to produce a text-to-speech system that offers an architecture that is easily adaptable to different languages and voices resulting in a significantly more natural voice than competing synthesizers, along with a talking head that enhances intelligibility and provides a rich multimodal communication experience.

## 2. TRAINING DATA

### 2.1 Lexical Data

We created a relational lexical database from three source lexica: The *Carnegie Mellon Pronouncing Dictionary* [16], *Moby Pronunciator II* [15] and *COMLEX English pronouncing lexicon* [8]. For speech synthesis, it is important to pronounce input orthographies correctly. We employ a stochastic disambiguator to tag incoming words for part of speech in order to select the appropriate pronunciations in the case of non-homophonous homographs. This requires that the lexicon contain part of speech tags for all words. The resulting database, Lexorola, contains almost 200,000 word entries, of which over 1500 are non-homophonous homographs. Sociolinguistic variants were removed in favor of one fairly consistent dialect. The idea behind removing sociolinguistic variation from the lexicon was to have a lexicon representing one plausible dialect of American English, and one from which various dialects and styles could be derived [9].

We refer to the level of transcription found in the dictionary as the "lexical" level. We contrast this with the "postlexical" level in accordance with lexical phonology [6]. Lexical pronunciations are characterized by their appropriateness for use in isolation, in contrast to postlexical pronunciations, which are appropriate in connected speech. We collect new lexical data for each major language or language variety we wish to synthesize. We can train a new letter-to-sound network for each such language or language variety, as described below.

### 2.2 Speech Data

The training vectors for most of the neural networks were derived from a single-speaker speech database. This database includes recordings of sentences from forty-eight of the Harvard sentence lists. The database has been augmented to include recordings of words spoken in isolation, questions, and paragraph-length recordings of texts from news stories.

These recordings have been hand-labeled on several levels. The phonetic labeling was performed in accordance with the procedure used for the TIMIT database [13]. Syllable, word, phrase, and clause boundaries were also marked. Each syllable was marked as having no stress, secondary stress or primary stress. Each word was tagged with a flag indicating whether it was a content or function word, and a number indicating expected prominence based on part of speech [11]. Finally, the rhythm and intonation of the speech was marked using the ToBI transcription system [1].

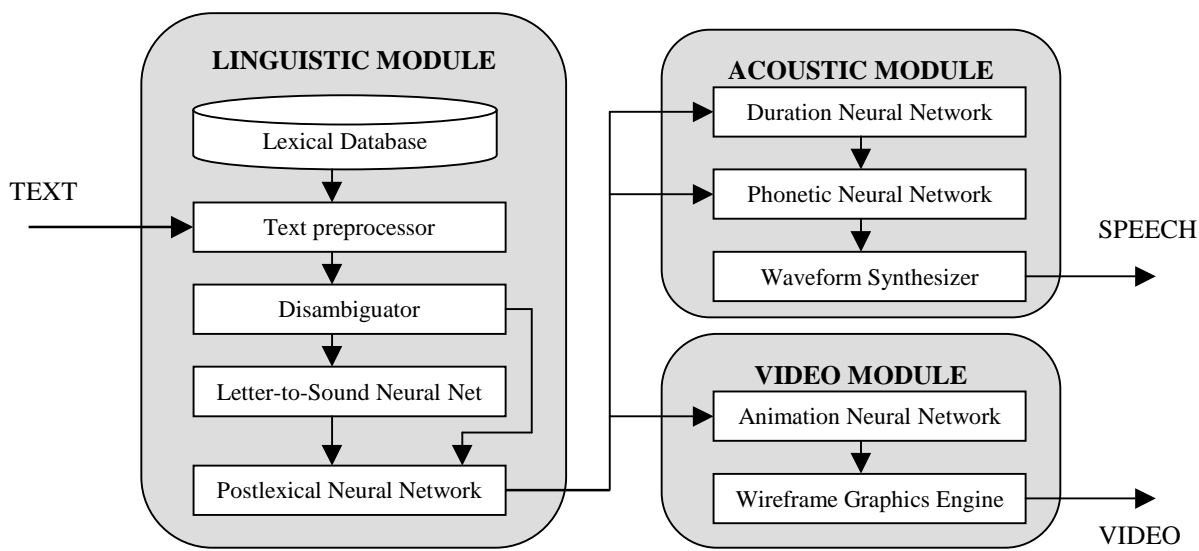

**Figure 1:** System Architecture

Information derived from this database was used to train the postlexical, duration, and phonetic neural networks described below. We record a new speech database for each voice to be synthesized by our system. We consequently retrain each of these neural networks for each new voice.

### 2.3 Video Data

In addition to the acoustic and linguistic data required to train the text-to-speech system, motion data must also be collected in order to generate an animated image. This animated image is preferably an accurate representation or characterized representation of the person whose voice is being modeled. The underlying model used to generate the animated image is a wireframe model.

The role of the video data is to allow the system to learn how the nodes on the wireframe model move as the subject speaks. In addition to the speech motions, gestures such as smiles, winks, eyebrow movements, and head nods are also recorded. Reference points, which correspond to the scaled wireframe model nodes, are located on the subject's face and black dots are placed at these locations. The marking of reference points is not necessary for the system but the motion tracking software which is used to extract the subject's movements can be greatly simplified by using physical marks placed on the subject. During the audio recording of the subject's voice, two video cameras are used to video tape the subject's motions from the front and the side angles.

## 3. SYSTEM OVERVIEW

As shown in Figure 1, the Motorola text-to-speech system consists of three principal modules: a linguistic module, an acoustic module and a visual module. The linguistic module is responsible for generating a linguistic representation from text. The acoustic module is responsible for generating speech from the linguistic representation. Finally, the visual module is responsible for driving a talking head based on the linguistic representation.

### 3.1 Linguistic Module

The linguistic module of the Motorola speech synthesizer is responsible for building a linguistic representation of user-supplied text to be spoken. To do this, we employ a text preprocessor that tokenizes textual input and looks up word pronunciations and tags in the lexical database. A stochastic disambiguator uses part of speech information to determine pronunciations in the case of non-homophonous homographs, such as *live* /lihv/ and *live* /layv/. At this point, syntactic and prosodic boundaries are also determined. Words that did not appear in the lexical database are submitted to a letter-to-sound neural network to determine their pronunciation. Finally, pronunciations from both the lexical database and the letter-to-sound neural network are submitted to a postlexical neural network that modifies pronunciations in a manner appropriate for the context in which they appear.

#### 3.1.1 Letter-to-sound neural network

Some words that are to be spoken will not be present in our lexicon, or any lexicon, no matter how large. These "out-of-dictionary" words may include some business, personal, and place names, neologisms and misspellings. Our approach is inspired by work using neural networks trained on dictionaries to determine pronunciations from orthography [12].

The first step in training the neural network to learn letter-phone correspondences is to align letters and phones in a meaningful way. In order to do this, we used a dynamic programming alignment algorithm [7]. While this algorithm is described for aligning sequences from the same alphabets, we needed to define a specialized cost function reflecting the likelihood that particular letters correspond with particular phones.

In addition to providing the network with aligned pronunciations and orthographies for each word in the dictionary, we provided the network with feature information for both phones and letters. We defined features for a letter to be the union of the features of the phones that that letter might represent. So the features for the letter 'c' would be the union of the features for /s/ and /k/.

While we are currently achieving competitive results, we believe that improved performance will come from simplifying the phonological representations found in the dictionary, for example, by removing allophones. We are confident that such simplifications will not detract from the quality of our ultimate output, due to the postlexical module, to be described next.

At this point, a preliminary linguistic representation of the utterance has been built from the input text. The linguistic representation organizes the utterance into a hierarchical prosodic phonological structure, including phones, syllables and phonological words, as well as higher order constituents.

### 3.1.2 Postlexical neural network

The linguistic representation is then submitted to a postlexical module where lexical pronunciations derived from the lexicon are converted to postlexical pronunciations typical of the speaker whose voice is being modeled. It is important to generate pronunciations typical of those upon which the acoustic neural networks were trained, in order to assure the highest quality speech output. In addition, learning the postlexical phenomena of each speaker to be synthesized assures that the speech will be natural and resemble the original talker as much as possible.

The modifications that take place in the postlexical module at present involve segmental insertions, deletions and substitutions, such as flapping and *t,d* deletion. In addition, the postlexical module learns any dialect differences between the lexical database and the voice database being modeled.

We aligned the speech database with lexical pronunciations from the lexicon used in letter-to-sound training. This alignment used the same dynamic programming algorithm as described above in the context of the letter-to-sound neural network's letter-phoneme alignment. A principal difference was in the substitution cost function employed. As described above, when sequences from different alphabets are to be aligned, it is important to explicitly define the cost of substituting particular members of those alphabets. In the case of the lexical-postlexical conversion, the alphabets are largely the same, but not identical. For example, phones like [q] and [dx] appear in the postlexical alphabet, but not in the lexical alphabet.

The neural network input coding employed a window of nine phones. The use of such a window allows for important contextual information to be accounted for by the net. For each phone, we provided feature information for both lexical and postlexical phones. In addition, for each phone, distance to word, phrase, clause, and sentence boundaries was included. In future work, we intend to make use of the prosodic hierarchy, provided by the database's ToBI annotation, in the neural network encoding.

In an unseen testing subset of the data, the lexical and postlexical phones were identical 70% of the time. That is, 30% of the time, there was a different postlexical phone from lexical phone. Testing of a neural network trained as described above on a segment of the database that was excluded from training resulted in 87% correct prediction of postlexical phones. This indicates that while substantial learning took place, there is still room for increased learning.

## 3.2 Acoustic Module

Once the text has been converted to a linguistic representation, the system converts the linguistic representation to speech in three stages. First, the timing of the speech signal is established by associating a segment duration with each phone in the linguistic representation. An acoustic representation, consisting of input parameters for the synthesis portion of a vocoder, is generated for each ten-millisecond frame of speech. Finally, the synthesis portion of the vocoder is used to generate speech from these acoustic descriptions.

### 3.2.1 Duration neural net

The neural network used to generate segment durations has been described in detail in a previous paper [2]. This network is trained to produce output representing the duration. Both the log of the segment duration and the number of standard deviations the duration differs from the mean duration for the phone have been used as output values. The input to the neural network identifies the phone, the stress on the syllable containing the phone, the type of word containing the phone, and which of the marked boundaries fall on the start and end of the phone. This information is contained in a context window describing the phone and several surrounding phones. In addition, the network has input describing the position of the phone relative to the nucleus of the syllable and the position of the syllable relative to prosodic boundaries and pitch accents. Finally, there is a bit vector input to the network describing which of a variety of conditions are met by the phone. These conditions were derived from a rule-based system for computing durations. Training vectors for this network were derived from the labels for the speech database described above.

### 3.2.2 Phonetic neural net

The phonetic neural network, which converts the linguistic representation and timing information, has also been described in previous papers [3][4][5]. The output of this neural network is the input to the synthesis section of the vocoder described in section 3.2.3. Training data is generated from the speech database using the analysis section of the vocoder. This is the most complicated neural network in the text-to-speech system, containing over 8000 inputs.

### 3.2.3 Vocoder

The phonetic neural network is not trained to generate speech directly. Instead, it is trained to produce a sequence of acoustic descriptions of ten-millisecond frames of speech. These are then synthesized using a vocoder. Since the neural network is not well suited to selecting entries from a codebook, a parametric vocoder was used. A mixed-mode vocoder, with excitation divided into a low-frequency voiced band and a high-frequency unvoiced band, was used. The parameters consist of the fundamental frequency, the power of the speech signal, the boundary frequency between the voiced and unvoiced bands, and ten line spectral frequencies. An analysis section converted the recorded speech from the database to these parameters for neural network training.

### 3.3 Video Module

The video subsystem takes the output of the linguistic module (section 3.1) and the output of the duration neural network (section 3.2.1) and generates an animated figure by using an additional neural network. The input to the neural network is exactly the same as the information used as input to phonetic neural network (section 3.2.2). The output is scaled coordinates of the nodes of the underlying wireframe model. In training the video neural network, the targets are the reference points that are extracted from the two video sequences that were recorded from a human subject (see section 2.3). Ideally these two video sequences are combined to generate 3-dimensional target points but system complexity can be reduced if a 2-dimensional wireframe model is used. During normal execution, the scaled output coordinates are used to control the shape of the wireframe model. Texture mapping is performed on the wireframe model to give the animated figure a natural appearance.

The use of a neural network in generating animated motion that is synchronized with the synthetic speech has several advantages. The first is that the system is able to automatically learn the idiosyncrasies of the specific person it was trained to imitate, which include characteristic movements and gestures. The high order equations contained within the neural network generate realistic movements, which lack the undesirable 'mechanical' or 'digital' qualities common in computer animation. The final major advantage to using neural network based animation is that the system is scaleable. The size of the neural networks can be reduced in order to meet system requirements while causing only a slight degradation in the performance of the video system.

## 4. PERFORMANCE

An independent speech perception lab conducted experiments comparing the acceptability and intelligibility of the phoneme-to-speech portion of the system to that of three commercial text-to-speech systems [10]. In the acceptability experiment, listeners were presented with sentences generated by the text-to-speech systems, and asked to rate them on a scale from one to seven, with one being unacceptable, and seven being highly acceptable. The results showed that the Motorola system was rated more acceptable than the other systems. In the intelligibility experiment, listeners were presented with words spoken in isolation, and asked to identify them. The Motorola systems did not do as well as some of the other systems in this experiment. At the time, the neural networks had not been trained using words spoken in isolation. Studies with networks trained on isolated words have not yet been carried out.

The system has been successfully implemented to operate in real time on both PowerPC and Pentium machines, running MacOS, Windows NT, and Windows 95.

## 5. SUMMARY

A text-to-speech system using neural networks for several of its components has been shown to be feasible. The speech produced from the system is found to be more acceptable to listeners than that of existing commercial systems. The incorporation of neural networks in multiple levels of a text-to-speech system permits rapid adaptability to new dialects and languages in comparison with other methods.

## 6. ACKNOWLEDGMENTS

The authors would like to thank Ira Gerson for his support throughout this research. We are grateful for the efforts of William Thompson, Peter Viechnicki and Erica Zeinfeld for their work on the creation and maintenance of the lexical database.